# Guess Who Rated This Movie: Identifying Users Through Subspace Clustering

**Amy Zhang**
Research Laboratory of Electronics
Massachusetts Institute of Technology
Cambridge, MA
amyzhang@mit.edu

**Nadia Fawaz, Stratis Ioannidis**
Technicolor
Palo Alto, CA
nadia.fawaz@technicolor.com
stratis.ioannidis@technicolor.com

**Andrea Montanari**
Departments of Statistics
and Electrical Engineering
Stanford University
Stanford, CA
montanari@stanford.edu

## Abstract

It is often the case that, within an online recommender system, multiple users share a common account. Can such shared accounts be identified *solely on the basis of the user-provided ratings*? Once a shared account is identified, can the different users sharing it be identified as well? Whenever such user identification is feasible, it opens the way to possible improvements in personalized recommendations, but also raises privacy concerns. We develop a model for composite accounts based on unions of linear subspaces, and use subspace clustering for carrying out the identification task. We show that a significant fraction of such accounts is identifiable in a reliable manner, and illustrate potential uses for personalized recommendation.

## 1 Introduction

Online commerce services such as Netflix provide personalized recommendations by collecting user ratings about a universe of items, to which we refer here as 'movies'. Typically, multiple people within a single household (family members, roommates, *etc.*) may share the same account for both viewing and rating movies. Service providers are avoid deploying multiple accounts as log-in screens are perceived as a nuisance and a barrier to using the service. This is especially true on a keyboard-less devices, such as televisions or gaming platforms. Account sharing persists even when providers offer the option of registering secondary accounts, as the latter may have access to a subset of the services enjoyed by the primary account. Finally, sharing might be regarded as a partial (if unconscious) privacy protection mechanism, hindering the release of the household's composition and demographics.

The use of a single account by multiple individuals poses a challenge in providing accurate personalized recommendations. Informally, the recommendations provided to a "composite" account, comprising the ratings of two dissimilar users, may not match the interests of either of these users. More concretely, as discussed in Section 3, collaborative filtering methods such as matrix factorization assume that ratings follow a linear model of user and movie profiles of small dimension. Though such methods may perform well for most cases, they can fail on composite accounts, as we show in Section 6. This is because "mixing" ratings from different users may yield a rating set that can no longer be explained by a linear model.

Can composite accounts within a recommender system be identified? Can the individuals sharing such an account be identified? Can accurate profiles of different users' behaviors be learnt? We address these questions in the most challenging setting, namely when no information is available apart from the ratings users provide. Our contributions are as follows:

(a) We develop a model of composite accounts as unions of linear subspaces. This allows us to apply a number of *linear subspace clustering* algorithms (Ma et al., 2008) to the present problem.

(b) Based on this model, we develop a statistical test that can be used as indicator of 'compositeness', and a model selection procedure to determine the number of users sharing the same account. We systematically apply and evaluate these methods on real datasets.

(c) In particular, we show that a significant fraction of composite accounts can be reliably identified. In a dataset made of both single-user and composite accounts, a subset $S$ of accounts can be selected that comprises roughly 70% of the composite accounts, while only 40% of the accounts in $S$ are single-user accounts.

(d) The users sharing an account can be identified

with good accuracy. For the accounts in the above set, more than 60% of the movies were identified correctly (a result that we estimate to be significant with $p < 0.05$).

(e) We apply these mechanisms on 54K Netflix users that rated more than 500 movies, and identify 4 072 composite users with high confidence.

(f) Finally, we demonstrate how the above methods can be applied to improve recommendations.

We consider this ability to identify multiple users behind an account quite surprising, in view that no information is used apart from users' ratings. In particular, *all publicly available datasets are susceptible to this identification*. Beyond personalized recommendations, this ability is useful/worrisome for a number of reasons. On one hand, it can aid in determining the household's demographics. Such information can be subsequently monetized, *e.g.*, through targeted advertising. On the other hand, user identification can be considered as a privacy breach, and calls for a careful privacy assessment of recommender systems.

The remainder of this paper is organized as follows. Section 2 briefly reviews related work. Section 3 develops our statistical model. Sections 4 and 5 apply and evaluate our new methods to recommender system datasets. Finally, Section 6 uses these methods to improve personalized recommendations.

## 2 Related Work

The problem of user identification from ratings has received attention only recently. The 2nd Challenge on Context-Aware Movie Recommendation (Said et al., 2011) addressed a "supervised" variant. Movie ratings generated by users in the same household as well as the ids of the users was provided as a training set. The test set included movie ratings attributed to households, and contestants were asked to predict which household members rated these movies. In contrast, we study an unsupervised version of the problem, where the mapping of movies to users is not a priori known.

To the best of our knowledge, we are the first to study user identification as a subspace clustering problem. Beyond EM and GPCA, several subspace clustering algorithms have been recently proposed (Elhamifar and Vidal, 2009; Liu et al., 2010; Soltanolkotabi and Candes, 2011; Eriksson et al., 2011). Preliminary simulations using these methods did not yield significant improvements.

## 3 Statistical Modeling

Consider a dataset of ratings on $M$ movies provided by $N$ accounts, each corresponding to a different household. Ratings are available for a subset of all $N \times M$ possible pairs: we denote by $\mathcal{M}_H \subseteq [M]$, where $m_H \equiv |\mathcal{M}_H|$, the set of movies rated by account/household $H$, and by $r_{Hj} \in \mathbb{R}$ the rating of movie $j \in \mathcal{M}_H$.

Each movie $j \in [M]$ is associated with a feature vector $\mathbf{v}_j \in \mathbb{R}^d$, where $d \ll N, M$. We use matrix factorization to extract the latent features for each movie, as described in Section 3.3. If explicit information (*e.g.*, genres or tags) is available, this can be easily incorporated in our model by extending the vectors $\mathbf{v}_j$.

Each household $H$ may comprise one *or more* users that actually rated the movies in $\mathcal{M}_H$. Abusing notation, we denote by $H$ the set of users in this household, and by $n_H = |H|$ the household size. For each $i \in H$, we denote by $A_i^* \subseteq \mathcal{M}_H$ the set of movies rated by $i$, and by $I^*(j) \in H$ the user that rated $j \in \mathcal{M}_H$.

Note that neither the household size $n_H$ nor the mapping $I^* : \mathcal{M}_H \to H$ are a priori known. We would like to perform the following inference tasks.

(a) *Model Selection*: determine the household size $n_H$. A closely related problem is the one of determining whether the account is *composite* (*i.e.*, $|H| > 1$) or not.

(b) *User Identification:* identify movies that have been viewed by the same user—*i.e.*, recover $I^*$, up to a permutation, and use this knowledge to profile the individual users.

We also explore the impact of user identification on targeted recommendations. The 'dual' impact on user privacy will be the object of a forthcoming publication.

### 3.1 Linear Model

We focus now on a single household, and omit the index $H$ hereafter. We thus denote by $n$ the household size, $\mathcal{M}$ and $m$ the set of movies rated by this household and its size, respectively, and by $r_j$ the rating given to movie $j \in \mathcal{M}$.

Our main modeling assumption is that the rating $r_j$ generated by a user $i \in H$ for a movie $j \in \mathcal{M}$ is determined by a linear model over the feature vector $\mathbf{v}_j$. That is, for each $i \in H$ there exists a vector $\mathbf{u}_i^* \in \mathbb{R}^d$ and a real number $z_i^* \in \mathbb{R}$ (the *bias*), such that

$$r_j = \langle \mathbf{u}_i^*, \mathbf{v}_j \rangle + z_i^* + \epsilon_j, \quad \text{for all } j \in A_i^*, i \in H, \quad (1)$$

where $\epsilon_j \in \mathbb{R}$ are i.i.d. Gaussian random variables with mean zero and variance $\sigma^2$. Such linear models are used extensively by rating prediction methods that rely on matrix factorization (Srebro and Jaakkola, 2003; Srebro et al., 2005; Koren et al., 2009), and are known to perform very well in practice.

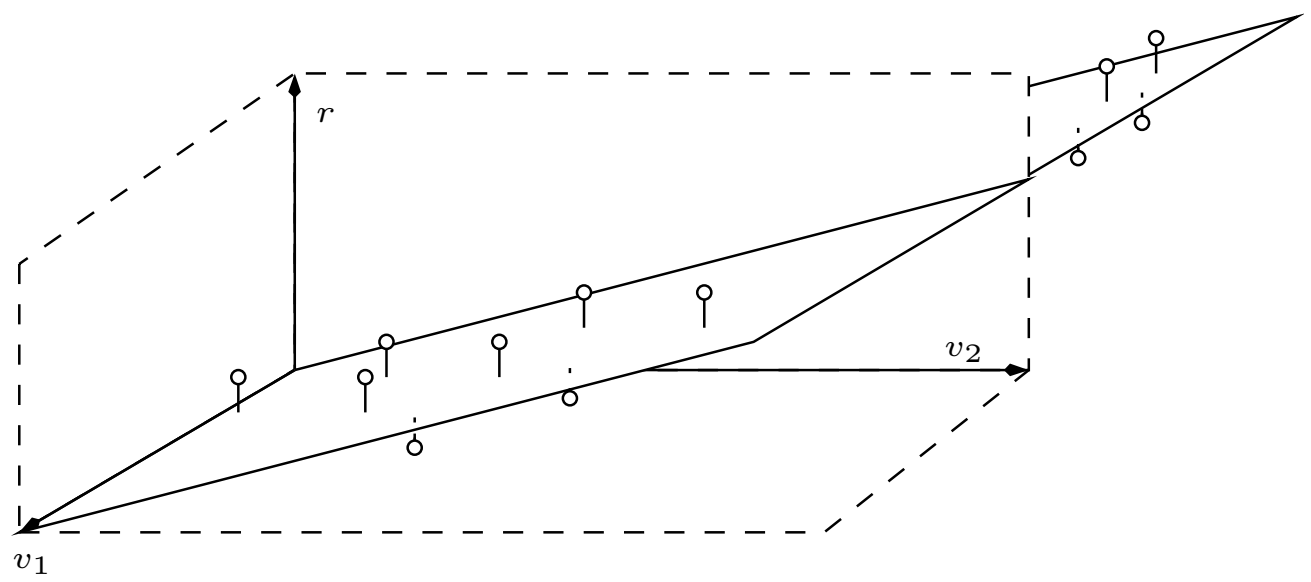

Figure 1: For all movies $j \in A_i$ rated by user $i \in H$, the points $\mathbf{x}_j = (\mathbf{v}_j, 1, r_j) \in \mathbb{R}^{d+2}$ lie slightly off a hyperplane whose normal is $(\mathbf{u}_i, z_i, -1) \in \mathbb{R}^{d+2}$.

Assuming that the household size is known, the model parameters of (1) are (a) the *user profiles* $\mathbf{\Theta}^* = \{\boldsymbol{\theta}_i^*\}_{i\in H} \in \mathbb{R}^{n\times d+1}$, where $\boldsymbol{\theta}_i^* = (\mathbf{u}_i^*, z_i^*) \in \mathbb{R}^{d+1}$, $i \in H$, as well as (b) the mapping $I^* : \mathcal{M} \to H$. Given two estimators $\mathbf{\Theta}, I$ of $\mathbf{\Theta}^*, I^*$, the log-likelihood of the observed sequence of pairs $\{(\mathbf{v}_j, r_j)\}_{j\in\mathcal{M}}$, is given by

$$L(\mathbf{\Theta}, I) = -\frac{1}{2\sigma^2}\sum_{j\in\mathcal{M}}\big(r_j - z_{I(j)} - \langle \mathbf{u}_{I(j)}, \mathbf{v}_j\rangle\big)^2. \quad (2)$$

Estimating the maximum likelihood model parameters thus amounts to minimizing the mean square error:

$$\min_{\mathbf{\Theta}, I}\ \mathrm{MSE}(\mathbf{\Theta}, I) = \frac{1}{m}\sum_{j\in\mathcal{M}}\big(r_j - z_{I(j)} - \langle \mathbf{u}_{I(j)}, \mathbf{v}_j\rangle\big)^2, \quad (3)$$

where $\mathbf{\Theta} \in \mathbb{R}^{n\times d+1}$, $I \in \mathcal{I}$, the set of all mappings from $\mathcal{M}$ to $H$. Note that (3) is not convex. Nevertheless, as discussed in Section 4.1, fixing $I$ results in a quadratic program, while fixing $\mathbf{\Theta}$ results in a combinatorial problem solvable in $O(nm)$ time.

### 3.2 Subspace Arrangements

We obtain an insightful geometric interpretation of the minimization (3) by studying the points $\mathbf{x}_j = (\mathbf{v}_j, 1, r_j) \in \mathbb{R}^{d+2}$, *i.e.*, the $d+2$-dimensional vectors resulting from appending $(1, r_j)$ to the movie profiles. Eq. (1) implies that although the points $x_j$ live in an ambient space of dimension $d+2$, they actually lie on a lower-dimensional manifold: the union of $n$ hyperplanes, *i.e.*, $d$+1-dimensional linear subspaces of $\mathbb{R}^{d+2}$.

To see this, let $\mathbf{n}_i^* = (\mathbf{u}_i, z_i, -1) \in \mathbb{R}^{d+2}$ be the vector obtained by appending the bias $z_i^*$ and -1 to $\mathbf{u}_i^*$. Then, $|\langle \mathbf{n}_i^*, x_j\rangle| = |\langle \mathbf{u}_i^*, \mathbf{v}_j\rangle + z_i^* - r_j| = |\epsilon_j|$, for every $j \in A_i$. Hence, provided that the variance $\sigma^2$ is small, the points $\mathbf{x}_j$ lie very close to the hyperplane with normal $\mathbf{n}_i^*$ that crosses the origin (see Figure 1).

A union of such affine subspaces is called a *subspace arrangement*. Given that the data $x_j$, $j \in \mathcal{M}$, "almost" lie on such a manifold, minimizing the MSE has the following appealing geometric interpretation. First, mapping a movie $j$ to a user amounts to identifying the hyperplane to which $x_j$ is closest to. Second, once movies are thus mapped to users, profiling a user amounts to computing the normal to its corresponding hyperplane. Finally, identifying the number of users in a household amounts to determining the number of hyperplanes in the arrangement.

These tasks are known collectively as the *subspace estimation* or *subspace clustering* problem, which has numerous applications in computer vision and image processing (Vidal, 2010). In Section 4.1, we exploit this connection to apply algorithms for subspace clustering on user identification (namely, EM and GPCA).

### 3.3 Datasets

We test our algorithms on two datasets:

**CAMRa2011 dataset.** The CAMRa2011 dataset was released at the Context-Aware Movie Recommendation (CAMRa) challenge at the 5th ACM International Conference on Recommender Systems (RecSys) 2011. This dataset consists of 4 536 891 5-star ratings provided by $N = 171\,670$ users on $M = 23\,974$ movies, as well as additional information about household membership for a subset of 602 users. The 290 households comprise 272, 14 and 4 households of size 2, 3 and 4 users, respectively. We use the entire dataset to compute the movie profiles $\mathbf{v}_j$ through matrix factorization, using $d = 10$ (found to be optimal through cross validation). In the sequel, we restrict our attention to the 544 users belonging to households of size 2. To simulate a composite account, we merge the ratings provided by users belonging to the same household. The original mapping of ratings to household members serves as the ground truth.

**Netflix Dataset.** The second dataset contains 5-star ratings given by $N = 480\,189$ users for $M = 17\,770$ movies. We again obtain the movie profiles $\mathbf{v}_j$ through matrix factorization on the entire dataset, with $d = 30$. We then restrict our attention to the subset of 54 404 users who rated at least 500 movies. We also generate 300 'synthetic' households of size 2 by pairing the ratings of 600 randomly selected users; we select these among the accounts that our model-selection methods, described in Section 5.3, classify as non-composite.

Matrix factorization is likely to be unreliable for extracting account feature vectors, as the latter may be composite. On the other hand, it appears to perform well for movies. We use the OptSpace algorithm of Keshavan et al. (2010) in both datasets for matrix factorization, which will not be further discussed.

# 4 User Identification

In this section, we address the user identification problem assuming that the household size $n$ is a priori known. This amounts to obtaining estimators of $I^*$ and $\boldsymbol{\theta}_i^* = (\mathbf{u}_i^*, z_i^*)$ for each user $i \in H$. We first describe four algorithms for solving this problem and then evaluate them on our two datasets. We present methods for determining the size $n$ in Section 5.

In the absense of any additional information, we cannot distinguish between two mappings $I : \mathcal{M} \to H$ that partition $\mathcal{M}$ identically. As such, we have no hope of identifying the correct "label" $i \in H$ of a user; we thus assume in the sequel, w.l.o.g., that $H = [1, \ldots, n]$.

## 4.1 Algorithms

**Clustering.** Our first approach consists of two steps. First, we obtain a mapping $I : \mathcal{M} \to [n] = H$ by clustering the rating events $(\mathbf{v}_j, r_j) \in \mathbb{R}^{d+1}$, $j \in \mathcal{M}$ into $n$ clusters. Second, given $I$, we estimate $\boldsymbol{\theta}_i = (\mathbf{u}_i, z_i)$, $i \in [n]$, by solving the quadratic program:

$$\min_{\boldsymbol{\Theta}} \mathrm{MSE}(\boldsymbol{\Theta}, I), \tag{4}$$

where MSE is given by (3). This is separable in each $\boldsymbol{\theta}_i$, so the latter can be obtained by solving

$$\min_{(\mathbf{u}_i, z_i)} \sum_{j \in A_i} (r_j - \langle \mathbf{u}_i, \mathbf{v}_j \rangle - z_i)^2. \tag{5}$$

where $A_i = \{j \in \mathcal{M} : I(j) = i\}$, which amounts to linear regression w.r.t. the model (1).

We perform the clustering in the first step using either (a) K-means or (b) spectral clustering. Each yields a distinct mapping; we denote the resulting two user identification algorithms by K-Means and Spectral, respectively. Intuitively, these methods treat the rating as "yet another" feature, and tend to attribute movies with very similar profiles $\mathbf{v}$ to the same user, even if they receive quite distinct ratings.

**Expectation Maximization.** The EM algorithm (Dempster et al., 1977) identifies the parameters of mixtures of distributions. It naturally applies to subspace clustering—technically, this is "hard" or "Viterbi" EM. Proceeding over multiple iterations, alternately minimizing the MSE in terms of the movie-user mapping $I$ and the user profiles $\boldsymbol{\Theta}$. Initially, a mapping $I^0 \in \mathcal{I}$ is selected uniformly at random; at step $k \geq 1$, the profiles and the mapping are computed as follows.

$$\boldsymbol{\Theta}^k = \operatorname*{arg\,min}_{\boldsymbol{\Theta} \in \mathbb{R}^{n \times (d+1)}} \mathrm{MSE}(\boldsymbol{\Theta}, I^{k-1}) \tag{6a}$$

$$I^k = \operatorname*{arg\,min}_{I \in \mathcal{I}} \mathrm{MSE}(\boldsymbol{\Theta}^k, I) \tag{6b}$$

The minimization in (6a) can be solved as in (4) through linear regression. Eq. (6b) amounts to identifying the profile that best predicts each rating, *i.e.*,

$$I^k(j) = \arg\min_{i \in H} (r_j - z_i^k - \langle \mathbf{u}_i^k, \mathbf{v}_j \rangle)^2, \quad j \in \mathcal{M}. \tag{7}$$

which can be computed in $O(nm)$ time.

**Generalized PCA.** The Generalized Principal Components Analysis (GPCA) algorithm, originally proposed by Vidal et al. (2005), is an algebraic-geometric algorithm for solving the general subspace clustering problem, as defined in section 3.2.

To give some insight on how GPCA works, we consider first an idealized case where the noise $\epsilon_j$ in the linear model (1) is zero. Then, the points $\mathbf{x}_j = (\mathbf{v}_j, 1, r_j)$, $j \in A_i^*$, lie exactly on a hyperplane with normal $\mathbf{n}_i^* = (\mathbf{u}_i^*, z_i^*, -1)$. Thus, every $\mathbf{x}_j$, $j \in \mathcal{M}$, is a root of the following homogeneous polynomial of degree $n$:

$$\begin{aligned} P_{\mathbf{c}}(\mathbf{x}) &= \prod_{i \in H} \langle \mathbf{n}_i^*, \mathbf{x} \rangle = \prod_{i \in H} \sum_{k=1}^{d+2} n_{ik}^* x_{jk} \\ &= \sum_{k_1 + \ldots + k_{d+2} = n, \forall l \ k_\ell \geq 0} c_{k_1, \ldots, k_{d+2}} x_1^{k_1} \ldots x_{d+2}^{k_{d+2}} \end{aligned} \tag{8}$$

We denote by $\mathbf{c} \in \mathbb{R}^{K(n,d)}$, where $K(n,d) = \binom{n+d+1}{n}$, the vector of the monomial coefficients $c_{k_1, \ldots, k_{d+2}}$. Note that $P_{\mathbf{c}}$ is uniquely determined by $\mathbf{c}$. Moreover, provided that $m = |\mathcal{M}| \geq K(n,d) = O(\min(n^d, d^n))$, $\mathbf{c}$ can be computed by solving the system of linear equations $P_{\mathbf{c}}(\mathbf{x}_j) = 0$, $j \in \mathcal{M}$.

Knowledge of $\mathbf{c}$ can be used to *exactly recover* $I^*$, up to a permutation. This is because, by (8), for any $j \in A_i^*$, the gradient $\nabla P_{\mathbf{c}}(x_j)$ is proportional to the normal $\mathbf{n}_i^*$. Hence, the partition in of points $\{A_i^*\}$ can be recovered by grouping together points with co-linear gradients (Vidal et al., 2005).

Unfortunately, this result does not readily generalize in the presence of noise (see, *e.g.*, Ma et al. (2008)). In this case, one approach is to estimate $\mathbf{p}$ by solving the (non-convex) optimization problem

$$\begin{aligned} \text{Minimize:} \ & \sum_{j \in [m]} ||\mathbf{x}_j - \widehat{\mathbf{x}}_j||_2^2 \\ \text{subject to:} \ & P_{\mathbf{c}}(\widehat{\mathbf{x}}_j) = 0 \end{aligned} \tag{9}$$

We use the heuristic of Ma et al. (2008) for solving (9) through a first order approximation of $P_{\mathbf{c}}$ and cluster gradients using the "voting" method also by Ma et al.

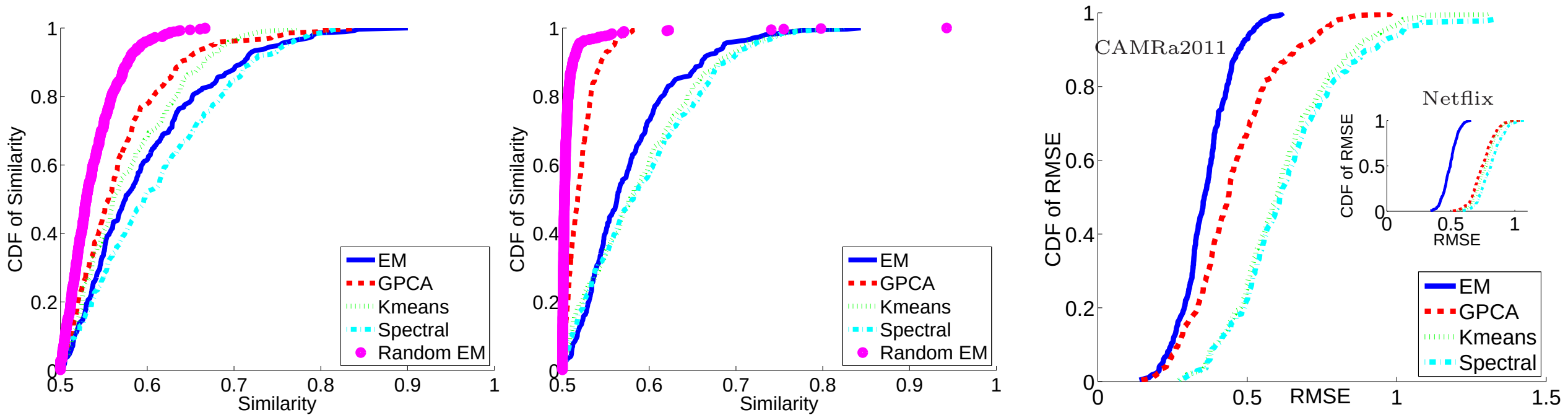

(a) CAMRa2011 Similarity (b) Netflix Similarity (c) CAMRa2011 and Netflix (inset) RMSE

Figure 2: Similarity and RMSE performance of K-Means, Spectral, EM, and GPCA, for households of size 2 in the CAMRa2011 and Netflix datasets.

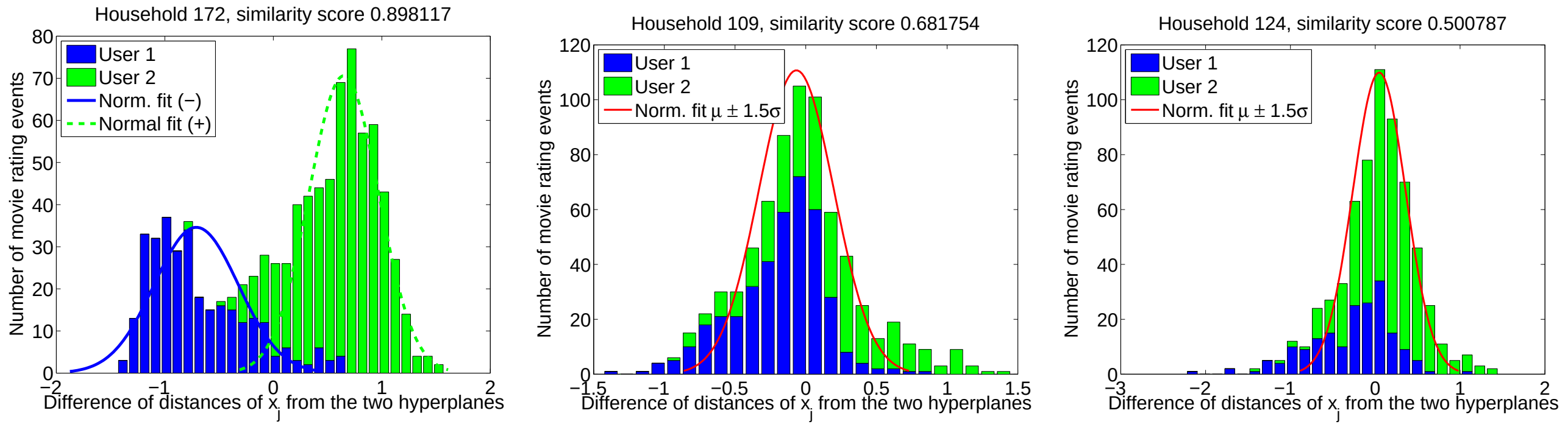

(a) High similarity (b) Intermediate similarity (c) Low similarity

Figure 3: PDF of the difference in distance of $\mathbf{x}_j$ from the two hyperplanes, computed by EM for three different households of size 2, ordered in decreasing similarity.

### 4.2 Evaluation

We evaluate the four algorithms, namely K-Means, Spectral, EM, and GPCA over CAMRa2011 and Netflix. In the CAMRa2011, we focus on the 272 composite accounts obtained by merging the ratings of users belonging to households of size 2. In Netflix, we focus on the 300 composite accounts obtained by pairing 600 users. For each composite account $H$, the original mapping $I^* : \mathcal{M} \to \{1, 2\}$ serves as ground truth.

**Similarity and RMSE.** We measure the performance of each algorithm two ways. First, we compare the mapping $I : \mathcal{M} \to \{1, 2\}$ obtained to the ground truth through the following *similarity* metric:

$$s(I, I^*) = \max_{\pi \in \Pi(\{1,2\})} \frac{1}{m} \sum_{j \in \mathcal{M}} \mathbb{1}\{\pi(I(j)) = I^*(j)\}$$

where $\Pi(\{1, 2\})$ is the set of permutations of $\{1, 2\}$. In other words, the similarity between $I$ and $I^*$ is the fraction of movies in $\mathcal{M}$ which $I$ and $I^*$ agree, up to a permutation. Notice that, by definition $s(I, I^*) \geq 0.5$. Second, we compute how well the obtained profiles $\boldsymbol{\Theta} = \{\boldsymbol{\theta}_i\}_{i \in H}$ fit the observed data by evaluating the root mean square error: $\text{RMSE}(\boldsymbol{\Theta}, I) = \sqrt{\text{MSE}(\boldsymbol{\Theta}, I)}$, where MSE is given by (3).

Figures 2a and 2b show the cumulative distribution function (CDF) of the similarity metric $s$ across all CAMRa2011 and Netflix composite accounts, respectively. Spectral performs the best in terms of similarity with EM (in CAMRa2011) and K-Means (in Netflix) being close seconds. The fact that clustering methods perform so well, in spite of treating ratings as "yet another" feature, suggests that users in these composite accounts indeed tend to watch different types of movies. Nevertheless, though K-Means and Spectral are comparable to EM in terms of $s$, they exhibit roughly double the RMSE of EM, as seen in Figure 2c. This is because, by grouping together similar movies with dissimilar ratings, these methods partition $\mathcal{M}$ in sets in which the linear regression (5) performs poorly.

**Statistical Significance.** In order to critically assess our results, we investigated the statistical significance of user identification performance under EM. We generated a null model by converting each of the 544 (600) users in the CAMRa2011 (Netflix) dataset into a composite account, by splitting the movies they rated into two random sets, thereby creating two fictitious users. Our random selection was such that the size ratio between the two sets in this partition followed the same distribution as the corresponding ratios in the real composite accounts. Our construction thus corresponds to a random "ground truth" that exhibits similar statistical properties as the original dataset. We subsequently ran EM over these 544 (600) fictitious composite accounts, and computed the similarity w.r.t. the random ground truth.

The resulting similarity metric CDF is indicated on Figures 2a and 2b as "Random EM". In CAMRa2011 (resp. Netflix), this curve indicates that any similarity $s > 0.59$ in CAMRa2011 (resp. $s > 0.52$) yields a p-value (probability of the similarity being larger or equal to $s$ under the null hypothesis) below 0.05. This corresponds to 41% and 88% of the composite accounts, respectively in each dataset. For these households, we can be confident that the high similarity performance is not due to random fluctuations.

**Precision at the Tail.** The similarity metric captures the performance of user identification in the aggregate across all movies in $\mathcal{M}$. Nevertheless, even when the similarity metric is extremely low, we can still attribute some movies to distinct users with very high confidence. As we will see in Section 5.3, this is important, because identifying even a few movies that a user has watched can be quite informative.

Let $(\mathbf{u}_i, z_i)$, $i \in \{1, 2\}$, be the profiles computed by EM for a given household $H$. Figure 3 shows histograms of the difference

$$\Delta_j = |r_j - \langle u_1, v_j \rangle - z_2| - |r_j - \langle u_2, v_j \rangle - z_2|, \quad (10)$$

for $j \in \mathcal{M}$, for three different composite accounts in CAMRa2011. Note that EM classifies $j$ as a movie rated by user 1 when $\Delta_j < 0$, and as a movie rated by user 2 otherwise. The three figures show the histograms of $\Delta_j$ for three households with high (0.90), intermediate (0.68), and low (0.50) similarity, respectively. The blue and green colors of each bar indicate the number of movies truly rated by user 1 and user 2, respectively. The total height of each bar corresponds to the total number of movies with that value of $\Delta_j$.

The household in Figure 3a exhibits a clear separation between the two users; indeed, $\Delta_j$ is negative for most movies rated by user 1 and positive otherwise. In contrast, in Figures 3b and 3c the distribution of $\Delta_j$ is concentrated at zero. Intuitively, a large number of movies are difficult to classify between users 1 and 2. Nevertheless, the tails of this distribution are overwhelmingly biased towards one of the two users. In other words, when labeling movies that lie on the tails of these distributions, our confidence is very high. We determine formally the tails of these curves by fitting a Gaussian on these histograms, after discarding points whose distance from the mean exceeds 1.5 standard deviations. Indeed, mapping the tails above these curves to distinct users identifies them accurately.

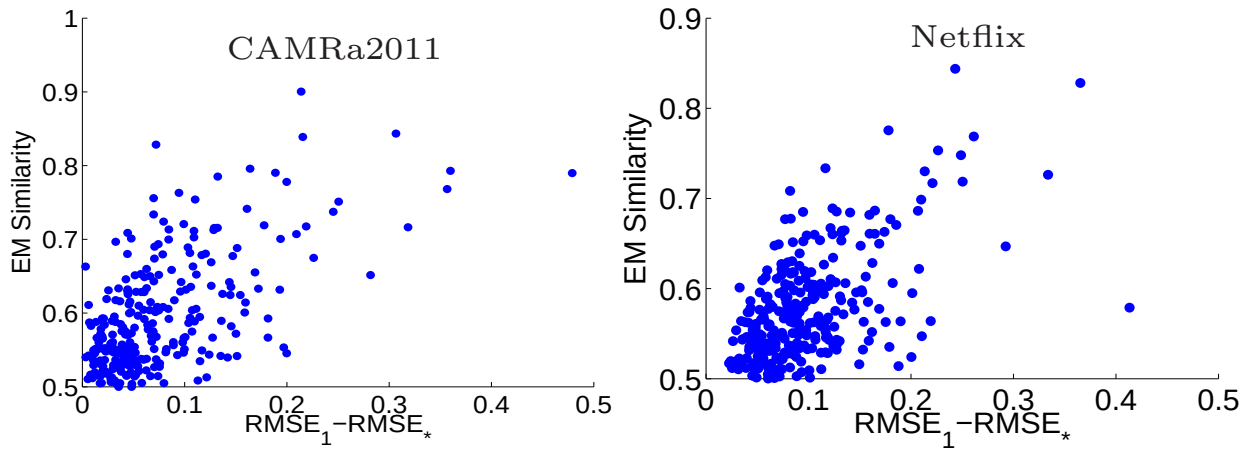

Figure 4: EM Similarity vs. gap in RMSE.

**Similarity Correlation to Diversity.** For each composite account, we computed the RMSE assuming that *all ratings were generated by a single user*: *i.e.*, we obtained a single profile $\boldsymbol{\theta}_1$ solving the regression (5), assuming that $I(j) = 1$ for all $j \in \mathcal{I}$, and used this to obtain an RMSE, denoted by $\text{RMSE}_1$. We also computed the RMSE assuming that *the mapping of ratings to users is known*: *i.e.*, we obtained two profiles $\boldsymbol{\theta}_1^*$ and $\boldsymbol{\theta}_2^*$ by solving the regression (5), assuming that $I = I^*$, and used these profiles to obtain a new RMSE, denoted by $\text{RMSE}_*$.

Figure 4 shows the similarity metric $s(I, I^*)$ for each composite account, computed using EM, versus the gap $\text{RMSE}_1 - \text{RMSE}_*$ for a particular household. We observe a clear correlation between the two values for both datasets. Intuitively, the EM method fails to identify users precisely on households where users *have similar profiles*, and for which distinguishing the users has little impact on the RMSE. The method performs well when users are quite distinct, and a single profile does not fit the observed data well.

## 5 Model Selection

The user identification methods presented in the previous section assume a priori knowledge of the number of users sharing a composite account. However, this information may not be readily available; in fact, determining if an account is composite or not is an interesting problem in itself. In this section, we propose and evaluate algorithms for this task.

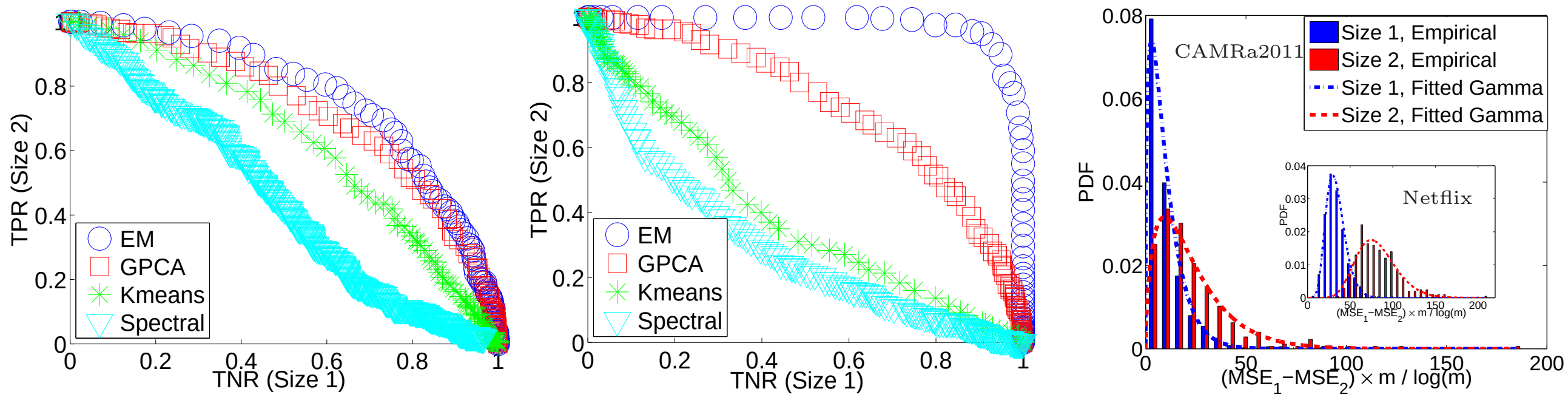

(a) CAMRa2011

(b) Netflix

(c) Normalized MSE Gap Distribution for CAMRa2011 and Netflix (inset).

Figure 5: (a)-(b): ROC curves, where TPR (TNR) is the number households correctly labeled as size two (one) over the number of households labeled as size two (one). (c): Normalized MSE gaps for sizes 1 and 2

### 5.1 Model Selection

The problem of estimating the number of unknown parameters in a model is known as *model selection*—see, *e.g.*, Hansen and Yu (2001). Denoting by $\mathbf{\Theta}_n \in \mathbb{R}^{n\times(d+1)}, I_n \in \mathcal{I}$ the estimators of the parameters $\mathbf{\Theta}^*, I^*$ of the linear model (1) for size $n$, the general method for model selection amounts to determining $n$ that minimizes $-\frac{1}{m}L(\mathbf{\Theta}_n, I_n) + \frac{C(\mathbf{\Theta}_n, I_n)}{m}$ where $L(\mathbf{\Theta}_n, I)$ is the log-likelihood of the data, given by (2), and $C$ is a metric capturing the *model complexity*, usually as a function of the number of parameters $n$. Several different approaches for defining $C$ exist; we report our results only for the *Bayesian Information Criterion* (BIC), by Schwarz (1978), as we observed that it performs best over our datasets.

The BIC for a household $H$ of size $|H| = n$ is given by

$$BIC_n := \frac{1}{2\sigma^2}\text{MSE}(\mathbf{\Theta}_n, I_n) + \frac{2n(d+1)\log m}{m}. \quad (11)$$

where $\sigma^2$ is the variance of the Gaussian noise in (1). Note that different methods for obtaining the estimators $\mathbf{\Theta}_n, I_n$ lead to different values for $BIC_n$.

We tested BIC on our two datasets as follows. For the CAMRa2011 (Netflix) dataset, we created a combined dataset comprising the 272 (300) composite accounts of $n = 2$ as well as as the 544 (600) individuals of size $n = 1$ that are included in these households, yielding a total of 816 (900) accounts. For each of these accounts, we first computed the MSE under the assumption that $n = 1$; this amounted to solving the regression 5 for a single profile $\boldsymbol{\theta}_1 = [\mathbf{u}_1, z_1]$ under $I(j) = 1$, for all $j \in \mathcal{M}$, obtaining an MSE we denote by $\text{MSE}_1$. Subsequently, we used each of the four identification methods (EM, GPCA, K-Means, and Spectral) to obtain a mapping $I : \mathcal{M} \to H$, and vectors $\boldsymbol{\theta}_i = (\mathbf{u}_i, z_i)$, $i \in \{1, 2\}$: each of these yielded an MSE for $n = 2$, denoted by $\text{MSE}_2$.

Using these values, we constructed the following classifier: we labeled an account as composite when

$$(MSE_1 - MSE_2) - \tau \log m/m > 0 \quad (12)$$

By varying $\tau$, we can make the classifier more or less conservative towards declaring accounts as composite. For $\tau = 2\sigma^2(d+2)$, this classifier coincides with BIC.

### 5.2 ROC Curves

The ROC curves obtained under different estimator functions for the model parameters can be found in Figure 5a for CAMRa2011. There is a clear ordering of the performance of different estimators as follows: EM (AUC=0.7711), GPCA (0.7455), K-Means (0.6111) and Spectral (0.4458). In particular, EM and GPCA yield very good classifiers. The performance on Netflix (Figure 5b) is even more striking, where EM (AUC=0.9796) significantly outperforms GPCA (0.7287), while K-Means (0.3879) and Spectral (0.2934) perform very poorly.

In Figure 5c, we plot the distribution of the normalized gap $(\text{MSE}_1 - \text{MSE}_2) \times m/\log m$ under EM for accounts of size 1 and 2, respectively. We see that in both datasets the distributions are well approximated by gamma distributions. Most importantly, accounts of size 2 exhibit a heavier tail. This is why labeling the outliers in the normalized gap distribution as households of size 2 as in (12) performs well.

### 5.3 Finding Composite Accounts on Netflix

**Model Selection in Netflix.** Armed with the above classification method, we turn our attention to the 54 390 users of the Netflix dataset that rated more than 500 movies. A natural question to ask is how many users in this dataset are in fact composite.

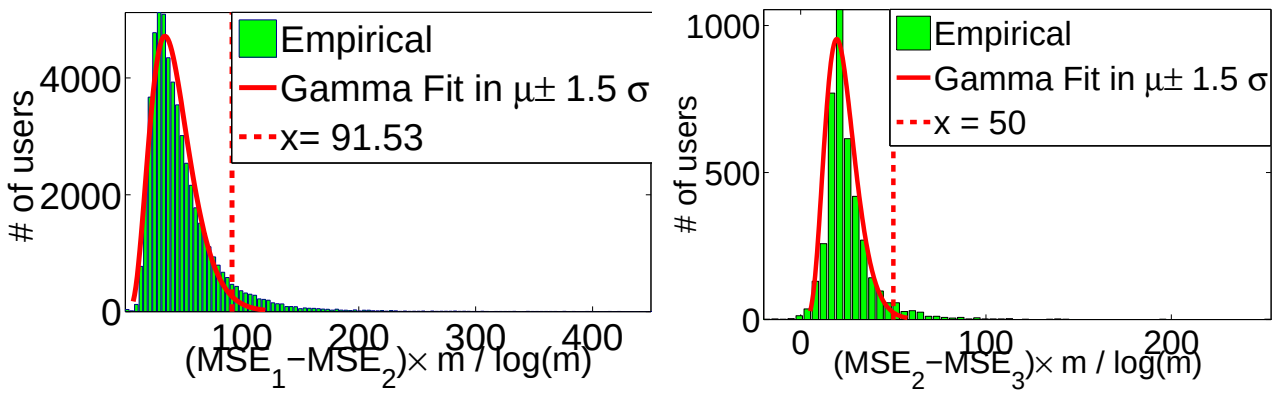

(a) Gap 1–2 in Netflix (b) Gap 2–3 in Netflix

Figure 6: PDF of RMSE gap in 54K Netflix users that rated more than 500 movies.

We first applied BIC with EM as a user identification method on these users. That is, we applied EM under the assumption the household size is $n$ =1,2, and 3, and labeled a household with the value $n$ that minimized $BIC_n$. We estimated the noise variance $\sigma^2$ through the mean square error of the matrix factorization applied on the entire dataset. The resulting classification labeled 36 832, 14 789, and 2 769 accounts as of size 1, 2, and $\geq 3$, respectively.

We also applied an alternative method (akin to the empirical Bayes approach of Efron (2009)). First, we plotted the histogram of the normalized gap in the MSE from a model of 1 to 2 users (Figure 6a). We then identified the outliers of this curve, and labeled them as accounts of size 2 and above. To identify the outliers, we fitted a gamma distribution to the portion of the histogram that lies within 1.5 standard deviations from the mean. Superimposing the two distributions, we found the normalized gap value (91.53) at which the tail of the original distribution (which has a heavier tail) had twice the value of the fitted gamma distribution; all accounts with a higher normalized gap were labeled as outliers. To identify accounts with size 3 and above, we repeated the above process only on the outliers, using now the normalized gap between models of 2 and 3 users (Figure 6b).

The resulting classification is compared to the classification under BIC in the following table:

| BIC\Outl. | 1 | 2 | $\geq 3$ | Tot. |
|---|---|---|---|---|
| 1 | 36832 | 0 | 0 | 36832 |
| 2 | 12712 | 2071 | 6 | 14789 |
| $\geq 3$ | 774 | 1805 | 190 | 2769 |
| Tot. | 50318 | 3876 | 196 | 54390 |

Note that the above method of outliers is more conservative when labeling accounts as composite, labeling only 4 072 users as composite. Nevertheless, we know that this method performs well over the datasets on which we have ground truth (c.f. Figure 5c).

**Visual Inspection.** Though we cannot assess the accuracy of this classification (we lack ground truth), a visual inspection of the accounts that were labeled as composite yield some interesting observations. Recall that, in each composite account, there are a few movies that we can assign to different users with very high confidence: these are precisely the movies that lie close to one of the two hyperplanes computed by EM and far from the other (*c.f.* Figure 3).

| User 1 | User 2 |
|---|---|
| TLOTR: The Fellowship of the Ring†(5), TLOTR: The Return of the King†(5), TLOTR: The Two Towers†(5), The Whole Nine Yards(4), Immortal†(1), The Deep End(2), Toys†(4), The Addams Family(5) | H.R. Pufnstuf(5), Sex and the City: Season 5♡(1), Me Myself & Irene(1), All the Real Girls♡△(5), Titanic♡(5), George Washington△(5), The Siege(1), In the Bedroom△(5) |

| User 1 | User 2 |
|---|---|
| Monsters Inc.◇(5), Finding Nemo◇(5), Whale Rider(5), Con Air(4), Lilo and Stitch◇(4), Ice Age◇(5), Ring of Fire(4), Star Trek: Nemesis(3), | In America♠(2), Super Size Me(2), A Very Long Engagement♠(1), Bend It Like Beckham(2), 21 Grams♠(1), Airplane II: The Sequel(4), Spun♠(1), Fahrenheit 9/11(1) |

Table 1: Movies rated by accounts labeled as composite in the Netflix dataset. We split each account using EM and show movies $j$ with most positive and most negative $\Delta_j$. Symbols indicate labels from the Netflix website: † = "Sci-Fi & Fantasy", ♡ = "Romantic", △ = "Understated", ◇ = "Children & Family Movies", ♠ = "Drama"

Using this intuition, we ran EM on several accounts declared as composite (size 2) by both BIC and the outlier method, and computed $\Delta_j$, given by (10) for each movie $j$ rated by these accounts. Table 1 shows the titles of the 8 most positive and 8 most negative movies for 2 such accounts. Looking up these titles on the Netflix website indicates clearly that these accounts exhibit a bimodal behavior. In the first account, 5/8 movies rated by User 1 are labelled as "Sci Fi & Fantasy", while 5/8 movies rated by User 2 are labelled either "Romantic" or "Understated". Similarly, in the second household, 4/8 movies rated by User 1 are labelled "Children & Family Movies", while 4/8 movies rated by User 2 are labelled as "Dramas", suggesting movies viewed by a child and an adult, respectively.

In many accounts we inspected, sequels (*e.g.*, "Lord of the Rings", "Star Wars", *etc.*) or seasons of the same TV show (*e.g.* "Sex and the City", "Friends", *etc.*) were grouped together (*i.e.*, attributed to the same user). The first account in Table 1 illustrates this.

We stress that we did not use any labeling or title information in our classification, as neither was available for both datasets. Nevertheless, as noted Section 3, such information can be incorporated in our model by extending $v_j$ to include any additional features.

## 6 Targeted Recommendations

In this section, we illustrate how knowledge of household composition can be used to improve recommen-dations. In a typical setup, a user accesses the account and the recommender system suggests a small set of movies from a catalog, recommending movies that are likely to be rated highly. However, even if the recommender knows the household composition and the user profiles, it still does not know who might be accessing the account at a given moment. In the absence of side information, we can circumvent this problem as follows. Assume the recommender has a budget of $K$ movies to be displayed; it can then recommend the union of the $K/n$ movies that are most likely to be rated highly by each of the $n$ users. This exploits household composition, without requiring knowledge of who is presently accessing the account.

To investigate the benefit of user identification, we performed a 5-fold cross validation in each of the 272 households in CAMRa2011, whereby user profiles where trained in 4/5ths of $\mathcal{M}$ (the training set), and used to predict ratings in the remaining 1/5th (the test set). As, in the real-life setting, we can circumvent identifying which user is accessing an account when recommending movies, we focus on predicting the ratings of users accurately. Ideally, we would like to assess our rating prediction over the test set for both users; unfortunately, we have the true rating of only one user for each movie. As a result, we assume that the mapping of movies to users is priori known on the test set (but *not* on the training set): to generate a prediction for a movie in the test set, we generate a single rating using the profile of the user that truly generated it.

We tested the following 4 methods. The first, termed *Single*, ignores the household composition; a unique profile $\boldsymbol{\theta}_S = (\mathbf{u}_i, z_i)$ is computed over the training set for both users through ridge regression over (5) using $I(j) = 1$ for all $j$ in the training set. The regularization parameter is chosen through cross validation. The second method, termed *Oracle*, assumes that the mapping of movies to users is known in the trainset; profiles $\theta_i^* = (\mathbf{u}_i^*, z_i^*)$, $i \in \{1, 2\}$ are obtained on the trainset using again ridge regression over (5) using $I = I^*$.

The third method, termed EM, uses the EM method outlined in Section 4 to obtain user profiles $\boldsymbol{\theta}_i = (\mathbf{u}_i, z_i)$. The EM algorithm is modified by adding regularization factor to the MSE, and using ridge rather than linear regression in each step. Finally, the last method, termed CNV for "convex", uses as a profile a linear combination of the common profile computed by Single and the specialized profile computed by EM. *I.e.* the profile of user $i$ is given by $\alpha\boldsymbol{\theta}_S + (1-\alpha)\boldsymbol{\theta}_i$, with $\alpha$ computed through cross validation.

We evaluate the performance of these methods in terms of two metrics. The first is the RMSE of the predicted ratings on the test set. The second, which we call *overlap*, is computed by generating a list of 6 movies and calculating the number of common elements with the 3 top rated movies by each user in the test set. For Single, the list is generated by picking the 6 movies in the test set with the highest predicted rating. For the remaining methods, we generate the list by picking the 3 movies in the test set with the highest predicted rating for each user, and combining these two lists.

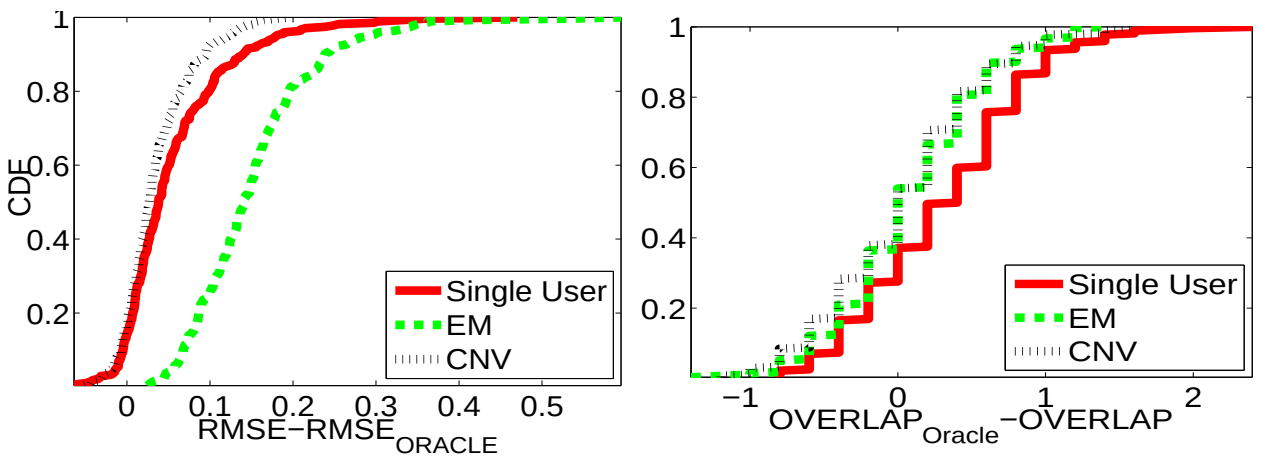

Figure 7: RMSE and OVERLAP performance compared to an oracle for the 272 composite users of CAMRa2011, when using (a) a single profile, (b) EM, and (c) a convex combination of the two.

Figure 7 shows the CDFs of the perfromance of the three mechanisms w.r.t. the distance of each metric from the corresponding metric under Oracle. We first observe that Oracle outperforms all other methods for the majority of the households, having an RMSE 0.60 and overlap 1.87, on average. This indicates that fitting a single profile to a composite account leads to poor predictions, which improve when the household composition is known. EM clearly outperforms Single w.r.t. the overlap metric, having a 14% higher overlap on average; however, it does worse w.r.t. RMSE also by roughly 14%. This is because, as observed in Figure 3, the bulk of movies are rated similarly by users, which dominates behavior in the RMSE; EM performs better on metrics that depend on the performance of outliers, such as overlap. In both metrics, CNV yields an improvement on both EM and Single, showing that the relative benefits of both methods can be combined.

## 7 Conclusion

We proposed methods for user identification solely on the ratings provided by users based on subspace clustering. Evaluating such methods in the presence of additional information is a potential future direction of this work. We also believe modeling rating data as a subspace arrangement can provide insight on a variety of applications, including privacy in recommender systems. In particular, altering or augmenting one's rating profile to *appear* as a composite user, with the purpose of obscuring, *e.g.*, one's gender, is an interesting research topic.